\documentclass[10pt,twocolumn,letterpaper]{article}

\usepackage{iccv}
\usepackage{times}
\usepackage{epsfig}
\usepackage{graphicx}
\usepackage{amsmath}
\usepackage{amssymb}
\usepackage{xfrac}
\usepackage[linesnumbered,algoruled,boxed,lined]{algorithm2e}

\usepackage[pagebackref=true,breaklinks=true,letterpaper=true,colorlinks,bookmarks=false]{hyperref}

 \iccvfinalcopy 

\def\iccvPaperID{5780} 

\ificcvfinal\fi
\begin{document}

\title{Event-based Face Detection and Tracking in the Blink of an Eye}

\author{Gregor Lenz, Sio-Hoi Ieng\\
Sorbonne Universit{\'e} INSERM, CNRS, Institut de la Vision\\
{\tt\small gregor.lenz@upmc.fr, sio-hoi.ieng@upmc.fr}
\and
Ryad Benosman\\
Sorbonne Universit{\'e}, University of Pittsburgh, Carnegie Mellon\\
{\tt\small benosman@pitt.edu}
}

\maketitle

\begin{abstract}
We present the first purely event-based method for face detection using the high temporal resolution of an event-based camera. We will rely on a new feature that has never been used for such a task that relies on detecting eye blinks. Eye blinks are a unique natural dynamic signature of human faces that is captured well by event-based sensors that rely on relative changes of luminance. Although an eye blink can be captured with conventional cameras, we will show that the dynamics of eye blinks combined with the fact that two eyes act simultaneously allows to derive a robust methodology for face detection at a low computational cost and high temporal resolution. We show that eye blinks have a unique temporal signature over time that can be easily detected by correlating the acquired local activity with a generic temporal model of eye blinks that has been generated from a wide population of users. We furthermore show that once the face is reliably detected it is possible to apply a probabilistic framework to track the spatial position of a face for each incoming event while updating the position of trackers. Results are shown for several indoor and outdoor experiments. We will also release an annotated data set that can be used for future work on the topic.
\end{abstract}

\begin{figure}[h]
	\centering
	\includegraphics[width=\columnwidth]{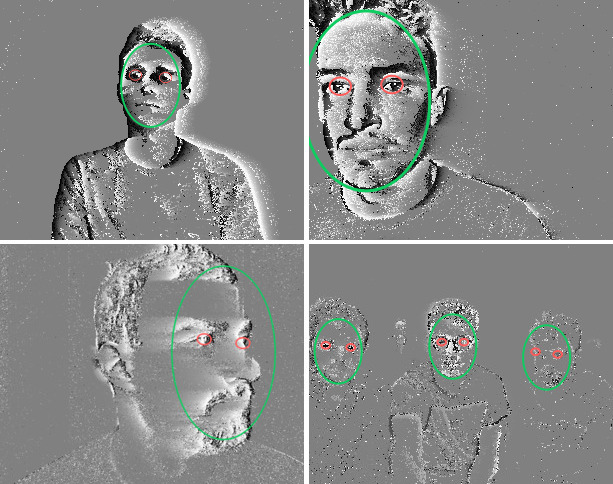}
	\caption{Event-based face tracking in different scenes. From left to right, top to bottom: \textbf{a}) indoors \textbf{b}) varying scale \textbf{c}) with one eye occluded \textbf{d}) multiple faces at the same time.}
	\label{fig:overview}
\end{figure}

\section{Introduction}
This paper introduces an event-based method to detect and track faces from the output of an event-based camera (samples are shown in Fig.\ref{fig:overview}). The method exploits the dynamic nature of human faces to detect, track and update multiple faces in an unknown scene. Although face detection and tracking is considered practically solved in classical computer vision, the use of conventional frame-based cameras does not allow to consider dynamic features of human faces. Event-based cameras record changes in illumination and are therefore able to record dynamics in a scene with high temporal resolution (in the range of 1$\mu$ to 1\,ms). In this work we will rely on eye blink detection to initialise the position of multiple trackers and reliably update their position over time. Blinks produce a unique space-time signature that is temporally stable across populations and can be reliably used to detect the position of eyes in an unknown scene. this paper extends the sate-of-art by:
\begin{itemize}
    \item implementing a low-power human eye-blink detection that exploits the high temporal precision provided by event-based cameras.
    \item detecting and tracking multiple faces simultaneously at $\mu$s precision.
\end{itemize}
The pipeline is entirely event-based in the sense that every event that is output from the camera is processed into an incremental, non-redundant scheme rather than creating frames from the events to recycle existing image-based methodology. We show that the method is inherently robust to scale change of faces by continuously inferring the scale from the distance of two eyes. Comparatively to existing image-based face detection techniques such as  ~\cite{ViolaRobustrealtimeface2004}\cite{JiangFaceDetectionFaster2017}\cite{liu2016ssd}, we show in this work that we can achieve a reliable detection at the native temporal resolution of the sensor without using costly computational techniques. Existing approaches usually need offline processing to build a spatial prior of what a human face should look like or vast amounts of data to be able to use machine learning techniques. 
The method is tested on a range of scenarios to show its robustness in different conditions: indoor and outdoor scenes to test for the change in lighting conditions; a scenario with a face moving close and moving away to test for the change in scale, a setup of varying pose and finally a scenario where multiple faces are detected and tracked simultaneously.
In order to compare performance to frame-based techniques, we build frames at a fixed frequency (25fps) from the grey-level events provided by the event based camera. We then apply gold-standard and state-of-the-art face detection algorithms on each frame and the results are used to assess the proposed event-based algorithm.

\subsection{Event-based cameras}
\begin{figure}[htb]
	\centering
	\includegraphics[width=\linewidth]{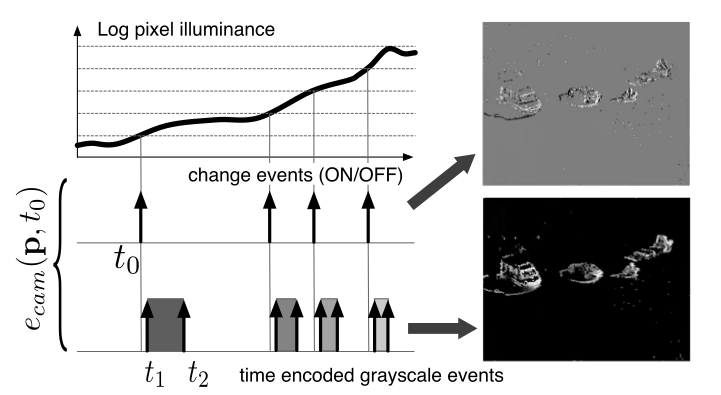}
	\caption{Working principle of the event-based camera and two types of events. 1) change event of type ON is generated at $ t_0$ as voltage generated by incoming light crosses a voltage threshold. 2) time $t_2 - t_1 $ to receive a certain amount of light is converted into an absolute grey-level value, emitted at $t_2$ used for ground truth in the paper.}
	\label{fig:ATIS}
\end{figure}
Event-based vision sensors are a new class of sensors based on an alternative signal acquisition paradigm. Rethinking the way how visual information is captured, they increasingly attract attention from  computer vision community as they provide many advantages that frame-based cameras are not able to provide without drastically increasing computational resources. Redundancy suppression and low latency are achieved via precise temporal and asynchronous level crossing sampling as opposed to the classical spatially dense sampling at fixed frequency implemented in standard cameras.

Most readily available event-based vision sensors stem from the Dynamic Vision Sensor (DVS)~\cite{LichtsteinerTemporalContrastVision2008}. As such, they work in a similar manner of capturing relative luminance changes. As Fig.~\ref{fig:ATIS} shows, each time illuminance for one pixel crosses a predefined threshold, the camera outputs what is called an event. An event contains the spatial address of the pixel, a timestamp and a positive (ON) or negative (OFF) polarity that corresponds to an increase or decrease in illuminance. Formally, such an event is defined as the n-tuple:
$ev=(x,y,t,p)$,
where $(x,y)$ are the pixel coordinates, $t$ the time of occurrence and $p$ is the polarity.
Variations of event-based cameras implement additional functionality.  In this work, we are using the Asynchronous Time-based Image Sensor (ATIS)~\cite{PoschQVGA143dB2011} as it also provides events that encode absolute luminance information, as does \cite{Guo2017}. Here the time it takes to reach a certain threshold is converted into an absolute grey-level value. This representation allows for easier comparisons with the frame-based world. To compare the output of such cameras with conventional ones, artificial frames can be created by binning the  grey-level events.  A hybrid solution of event- and frame-based world captures grey-level frames like a regular camera on top of the events~\cite{brandli2014240}. 
Inherently, no redundant information is captured, which results in significantly lower power consumption. The amount of generated events directly depends on the activity and lighting conditions of the scene. Due to the asynchronous nature and therefore decoupled exposure times of each pixel these sensors timestamp and output events with $\mu$s precision and are able to reach a dynamic range of up to 125\,dB. The method we propose can be applied to any event-based camera operating at sub-millisecond temporal precision as it only uses events that encode change information.



\subsection{Face detection}
The advent of neural networks enables state-of-the-art object detection networks that can be trained on facial images \cite{YangFacenessNetFaceDetection2017, JiangFaceDetectionFaster2017, sun2018face}, which rely on intensive computation of static images and need enormous amounts of data. Although there have been brought forward ideas on how to optimise frame-based techniques for face detection on power-constraint phones, most of the times they have to use a dedicated hardware co-processor to enable real-time operation~\cite{ren2008real}. Nowadays dedicated chips such as Google's Tensor Processing Unit or Apple's Neural Engine have become an essential part in frame-based vision, specialising in executing the matrix multiplications necessary to infer neural networks on each frame as fast as possible. In terms of power efficiency algorithms such as the one developed by Viola and Jones \cite{ViolaRobustrealtimeface2004} are still more than competitive. 

Dedicated blink detection in a frame-based representation is a sequence of detections for each frame. To constrain the region of interest, a face detection algorithm normally is used beforehand. Blinks are then deduced from the coarse sequence of detection results, which depending on the frame rate typically ranges from 15 to 25\,Hz\cite{NomanMobileBasedEyeBlinkDetection2018}. In an event-based approach, we turn the principle inside out and use blink detection as a mechanism to drive the face detection and tracking. Being the first real-time event-based face detector and tracker (to the best of our knowledge), we show  that by manually labelling fewer than 50 blinks, we can generate sufficiently robust models that can be applied to different scenarios. The results clearly contrast the vast amount of data and GPUs needed to train a neural network. 

\subsection{Human eye blinks}
We take advantage of the fact that adults blink synchronously and more often than required to keep the surface of the eye hydrated and lubricated. The reason for this is not entirely clear, research suggests that blinks are actively involved in the release of attention \cite{NakanoBlinkrelatedmomentaryactivation2013}. Generally, observed eye blinking rates in adults depend on the subject's activity and level of focus and can range from $3\,\sfrac{blinks}{min}$ when reading up to $30\,\sfrac{blinks}{min}$ during conversation (Table \ref{table_mean_blinking_rates}). Fatigue significantly influences blinking behaviour, increasing both rate and duration \cite{JohnA.SternBlinkRatePossible1994}. 
Typical blink duration is between $100 - 150\,ms$ \cite{BenedettoDriverworkloadeye2011} and shortens with increasing physical workload or increased focus. 

\begin{table}[h]
	\renewcommand{\arraystretch}{1.3}
	\centering
	\setlength{\tabcolsep}{9pt}
	\begin{tabular}{r|p{1cm}|p{1cm}}
		Activity & \multicolumn{2}{c}{\# Blinks per min} \\
		\hline
		reading & 4.5 & 3-7\\
		at rest  & 17 &-  \\
		communicating& 26 & - \\
		non-reading &- & 15-30 \\
	\end{tabular}
	\caption{Mean blinking rates according to \cite{BentivoglioAnalysisblinkrate1997} (left column) and \cite{JohnA.SternBlinkRatePossible1994} (right column)
	\label{table_mean_blinking_rates}}
\end{table}

\begin{figure}[htb]
	\centering
	\includegraphics[width=\linewidth]{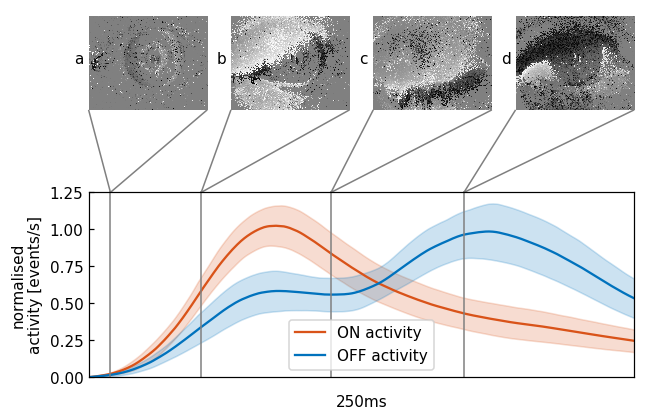}
	\caption{Mean and variance of the continuous activity profile of averaged blinks in the outdoor data set with a decay constant of 50\,ms. a) minimal movement of the pupil, almost no change is recorded. b) eye lid is closing within 100\,ms, lots of ON-events (in white) are generated. c) eye is in a closed state and a minimum of events is generated. d) opening of the eye lid is accompanied by the generation of mainly OFF-events (in black).}
	\label{fig:single-blink}
\end{figure}

To illustrate what happens during an event-based recording of an eye blink, Fig. \ref{fig:single-blink} shows different stages of the eye lid closure and opening. If the eye is in a static state, few events will be generated (a). The closure of the eye lid happens within 100\,ms and generates a substantial amount of ON events, followed by a slower opening of the eye (c,d) and the generation of mainly OFF events. From this observation, we devise a method to build a temporal signature of a blink. This signature is then used to signal the presence of a pair of eyes in the field of view, hence the presence of a face.

\section{Methods} 
\subsection{Temporal signature of an eye blink} 
\begin{figure*}[t]
	\centering
	\includegraphics[width=\linewidth]{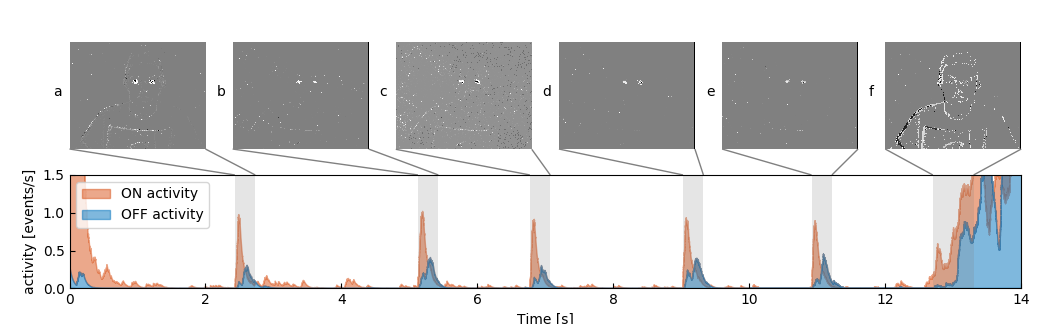}
	\caption{Showing ON (red) and OFF (blue) activity for one tile which lines up with one of the subject's eyes. Multiple snapshots of accumulated events for 250\,ms are shown, which corresponds to the grey areas.\textbf{a-e)} Blinks. Subject is blinking. \textbf{f)} Subject moves as a whole and a relatively high number of events is generated.}
	\label{fig:activity-timeline}
\end{figure*}

Eye blinks are a natural dynamic stimulus that can be represented as a temporal signature. While a conventional camera is not adequate to produce such a temporal signature because of its stroboscopic and slow acquisition principle, event-based sensors on the contrary are ideal for such a task. The blinks captured by an event-based camera are patterns of events that possess invariance in time because the duration of a blink is independent of lighting conditions and steady across the population. To build a canonical eye blink signature $A(t_i)$ of a blink, we convert events acquired from the sensor into temporal activity. For each incoming event $ev=(x_i,y_i,t_i,p_i)$, we update $A(t_i)$ as follows:
\begin{equation}
A(t_i)=
\left\{
\begin{matrix}
A_{on}(t_{u})  e^{-\frac{t_i-t_{u}}{\tau}} + \frac{1}{scale} & \textrm{if $p_i$=ON}\\
A_{off}(t_{v})  e^{-\frac{t_i-t_{v}}{\tau}} + \frac{1}{scale} & \textrm{if $p_i$=OFF}
\end{matrix}
\right.
\label{eq:activity_increase}
\end{equation}
where $t_u$ and $t_v$ are the times an ON or OFF event occurred before $t_i$. The respective activity function is increased by $\frac{1}{scale}$ at each time $t_n$ an event ON or OFF is registered. The quantity $\textrm{scale}$ acts as a corrective factor to account for a possible change in scale, as a face that is closer to the camera will inevitably trigger more events. Fig.~\ref{fig:activity-timeline} on top of the next page shows the two activity profiles for one tile that aligns with the subject's eye in a recording. Clearly visible are the 5 profiles of the subject's blinks, as well as much higher activities at the beginning and the end of the sequence when the subject moves as a whole. From a set of manually annotated blinks we build such an activity model function as shown in Fig.~\ref{fig:single-blink} where red and blue curve respectively represent the ON and OFF parts of the profile.

Our algorithm detects blinks by checking whether the combination of local ON- and OFF-activities correlates with that model blink that had previously been built from annotated material. 
To compute that local activity, the overall input focal plane is divided into one grid of 16 by 16 tiles, overlapped with a second similar grid made of 15 by 15 tiles. Each of these are rectangular patches of $19 \times 15$ pixels, given the event-camera's resolution of $304 \times 240$ pixels. They have been experimentally set to line up well with the eyes natural shape. The second grid is shifted by half the tile width and height to allow for redundant coverage of the focal plane. 

An activity filter is applied to reduce noise:
For each incoming event, its spatio-temporal neighbourhood is checked for corresponding events. If there are no other events within a limited time or pixel range, the event is discarded. Events that pass this filter will update ON or OFF activity in their respective tile(s) according to Eq.~\ref{eq:activity_increase}. Due to the asynchronous nature of the camera, activities in the different tiles can change independently from each other, depending on the observed scene.


\subsubsection{Blink model generation}
The model blink is built from manually annotated blinks from multiple subjects. We used two different models for indoor and outdoor scenes, as the ratio between ON and OFF events changes sufficiently in natural lighting. 20 blinks from 4 subjects resulted in an average model as can be seen in Fig.~\ref{fig:single-blink}. The very centre of the eye is annotated and events within a spatio-temporal window of one tile size and 250\,ms are taken into account to generate the activity for the model. This location does not necessarily line up with a tile of the previously mentioned grids. Due to the sparse nature of events, we might observe a similar envelope of activity for different blinks, however the timestamps of when events are received will not be exactly the same. Since we want to obtain a regularly sampled, continuous model, we interpolate activity between events by applying Eq.~\ref{eq:activity_increase} for a given temporal resolution  $R_t = 100 \mu s$. Those continuous representations for ON (red curve) and OFF (blue curve) activity are then averaged across different blinks and smoothed to build the model. Grey area in Fig.~\ref{fig:sparse-correlation} representing such a continuous model corresponds to blue mean in Fig.~\ref{fig:single-blink}. We define the so obtained time-continuous model:
\begin{equation}
    B(t)=B_{ON}(t) \cup B_{OFF}(t).
\end{equation}
As part of the model and for implementation purposes, we are also adding the information $N= \frac{\#\textrm{events}}{T.scale}$, which is normalised by the scale term that reflects the typical number of events triggered by a blink within that last $T$ms. 

\subsubsection{Sparse cross-correlation} 
\begin{figure}[htb]
\begin{center}
\includegraphics[width=\columnwidth]{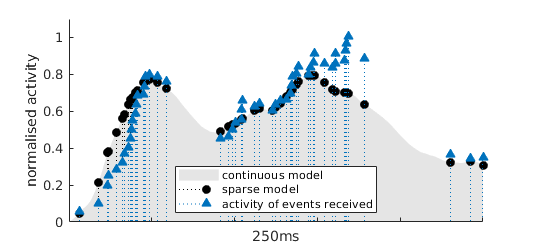}
\caption{
Example of a sparse correlation for OFF activity of an actual blink. The grey patch represents $B_{OFF}$, the activity model for OFF events previously built for outdoor data sets. Blue triangles correspond to the activity $A(t_k)$ for which events have been received in the current time window. Black dots symbolise $B_{OFF}(t_k)$, the value of activity in the model at the same times-tamps as incoming events. Values for blue triangles and black dots will be correlated to obtain the similarity score.
\label{fig:sparse-correlation}}
\end{center}
\end{figure}
When streaming data from the camera, the most recent activity within a $T=250$\,ms time window is taken into account in each tile to calculate the template matching score for ON and OFF activity. 
However, the correlation score is only ever computed if the number of recent events exceeds $N$, to avoid costly and unnecessary calculations. To further alleviate computational burden, we harness the event-based nature of the recording by taking into account only values for which we have received events. Fig.\,\ref{fig:sparse-correlation} shows an example of a sparse correlation calculation. The cross-correlation score between the incoming stream of events and the model is given by:
\begin{equation}
    C(t_k)=\alpha C_{on}(t_k)+(1-\alpha)C_{off}(t_k),
    \label{eq:corr1}
\end{equation}
where 
\begin{equation}
    C_p(t_k) = \displaystyle\sum_{i=0}^{N} A_{p}(t_i)B_{p}(t_i-t_k),
    \label{eq:corr2}
\end{equation}
with $p\in\{ON,OFF\}$. The ON and OFF parts of the correlation score are weighted by a parameter $\alpha$ that tunes the contribution of the ON/OFF events. This is necessary as, due to lighting and camera biases, ON and OFF events are usually not balanced. The weight $\alpha$ is set experimentally, typically for indoor and outdoor conditions. 

It is important for implementation reason, to calculate the correlation as it is in Eq.~\ref{eq:corr2} because while it is possible to calculate the value of the model $B(t_m-t_k)$ at anytime, samples for $A$ are only known for the set of times $\{t_i\}$, from the events.

If $C(t_i)$ exceeds a certain threshold, we create what we call a blink candidate event for the tile in which the event that triggered the correlation occurred. Such a candidate is represented as the n-tuple $eb = (r, c, t)$, where $(r,c)$ are the coordinates of the grid tile and $t$ is the timestamp. We do this since we correlate activity for tiles individually and only in a next step combine possible candidates to a blink. 

\subsubsection{Blink detection}
To detect the synchronous blinks of two eyes, blink candidates across grids generated by the cross-correlation are tested against additional constraints for verification. As a human blink has certain physiological constraints in terms of timing, we check for temporal and spatial coherence of candidates in order to find true positives. The maximum temporal difference between candidates will be denoted as $\Delta T_{max}$ and is typically 50\,ms, the maximum horizontal spatial disparity $\Delta H_{max}$ is set to 60\,pixels and maximum vertical difference $\Delta V_{max}$ is set to 20\,pixels. Algorithm~\ref{algo:blink-detection-pseudo-code} summarises the set of constraints to validate a blink. We trigger this check whenever a new candidate is stored. The scale factor here refers to a face that has already been detected.

\begin{algorithm}[htb]
\caption{Blink detection}

\label{algo:blink-detection-pseudo-code}
    \textbf{Inputs:} A pair of consecutive blink candidate events $eb_u=(r_u,c_u,t_u)$ and $eb_v=(r_v,c_v,t_v)$ with $t_u > t_v$ \\
    \If{ ($t_{u} - t_{v} < \Delta T_{max}$) AND ($|r_{u} - r_{v}| < \Delta V_{max} \times scale$) AND ($|c_{c} - c_{v}| < \Delta H_{max} \times scale$)}{
    \eIf{face is a new face}{\textbf{return} 2 trackers with $scale = 1$}{\textbf{return} 2 trackers with previous $scale$}}{}
 \end{algorithm}
 
\subsection{Gaussian tracker}
Once a blink is detected with sufficient confidence, a tracker is initiated at each detected location.
Trackers such as the ones presented in~\cite{LagorceAsynchronousEventBasedMultikernel2015} are used with bivariate normal distributions to  locally model the spatial distribution of events. For each event, every tracker is assigned a score that represents the probability of the event being generated by the tracker: 
\begin{equation} \label{eq:tracker_probability}
p(u) = \frac{1}{2\pi}|\Sigma|^{-\frac{1}{2}} e^{-\frac{1}{2}(\mathbf{u-\mu})^T \Sigma^{-1}(\mathbf{u-\mu})}
\end{equation}
where $ u= [x,y]^T$ is the pixel location of the event and the covariance matrix $\Sigma$ is determined when the tracker is initiated and will also update according to the distance between the eyes. The tracker with the highest probability is updated, provided that it is higher than a specific threshold value. A circular bounding box for the face is drawn based on the horizontal distance between the two eye trackers. We shift the centre of the face bounding box by a third of the distance between the eyes to properly align it with the actual face. 

\subsection{Global algorithm}
The detection and tracking blocks put together allow us to achieve the following event-by-event global face tracking Algorithm~\ref{algo:general_pseudo_code}: 
\begin{algorithm}[htb]
\caption{Event-based face detection and tracking algorithm} \label{algo:general_pseudo_code}
	\For{each event ev(x, y, t, p)}{
		\If{at least one face has been detected}{
		update best blob tracker for $ev$ as in (\ref{eq:tracker_probability})
		
		update $scale$ of face for which tracker has moved according to tracker distance
		}{
		update activity according to (\ref{eq:activity_increase})
		
		correlate activity with model blink as in (\ref{eq:corr1})
		
		run Algorithm~\ref{algo:blink-detection-pseudo-code} to check for a blink
        }
	}
\end{algorithm}	

\section{Experiments and Results}
We evaluated the algorithm's performance on a total of 48 recordings from 10 different people. The recordings are divided into 4 sets of experiments to assess the method's aptitude under realistic constraints encountered in natural scenarios. The event-based camera is static, observing people interacting or going after their own tasks. The data set tested in this work includes the following parts:
\begin{itemize}
    \item a set of indoor sequences showing individuals moving in front of the camera.
    \item a set of outdoor sequences similarly showing individuals moving in front of the camera.
    \item a set of sequences showing a single face moving back and forth w.r.t. the camera to test for scale change robustness.
    \item a set of sequences with several people discussing, facing the camera to test for multi-detections.
    \item a set of sequences with a single face changing its orientation w.r.t. the camera to test for occlusion resilience.
\end{itemize}
The presented algorithm has been implemented in C++ and runs in real-time on an Intel Core i5-7200U CPU. We are quantitatively assessing the proposed method's accuracy by comparing it with state of the art and gold standard face detection algorithms from frame-based computer-vision. As these approaches require frames, we are generating grey-levels from the camera when this mode is available. The Viola-Jones~\cite{ViolaRobustrealtimeface2004} algorithm provides the gold standard face detector and a Faster R-CNN and a Single Shot Detector (SSD) network that have been trained on the Wider Face\cite{YangWiderfaceface2016} data set enable comparison with state-of-the-art face detectors based on deep learning~\cite{ren2015faster, liu2016ssd}.


\subsection{Blink detection and face tracking}
The proposed blink detection and face tracking technique requires reliable detections or true positives. We do not actually need to detect all blinks because one is already sufficient to initiate the trackers. Additional incoming blink detections are used to correct trackers' drifts from time to time and could possibly decrease latency until tracking starts. As we will show in the experimental results, blinks are usually detected with a ratio of 60\% which ensures reliable tracking accuracy.

\subsection{Indoor and outdoor face detection}
\begin{figure}[ht]
	\centering
	\includegraphics[width=\columnwidth]{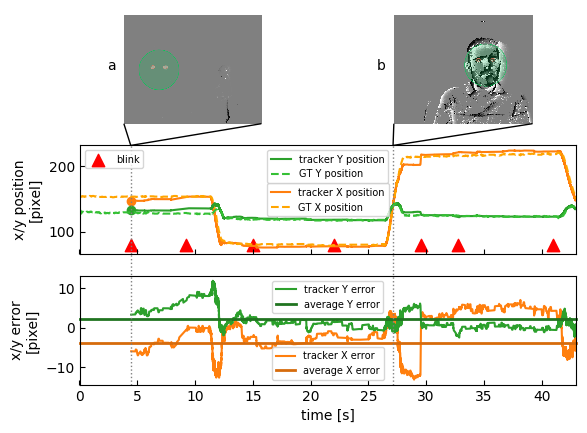}
	\caption{face tracking of one subject over 45s. a) subject stays still and eyes are being detected. Movement in the background to the right does not disrupt detection. b) when the subject moves, several events are generated}
	\label{fig:whole-data-recording}
\end{figure}

The indoor data set consists of recordings in controlled lighting conditions. As blinking rates are highest during rest or conversation, subjects in a chair in front of the camera were instructed not to focus on anything in particular and to gaze into a general direction. Fig.~\ref{fig:whole-data-recording} shows tracking data for such a recording. Our algorithm starts tracking as soon as one blink is registered (a). After an initial count to 10, the subject should lean from side to side every 10 seconds in order to vary their face's position. Whereas tracking accuracy on the frame-based implementation is constant (25\,fps), our algorithm is updated event-by-event depending on the movements in the scene. If the subject stays still, computation is drastically reduced as there is a significantly lower number of events. Head movement causes the tracker to update within $\mu s$ (b), incrementally changing its location in sub-pixel range. Eye trackers that go astray will be rectified at the next blink.\\

Subjects in the outdoor experiments were asked to step from side to side in front of a camera placed in a courtyard under natural lighting conditions. Again they were asked to gaze into a general direction, partly engaged in a conversation with the person who recorded the video. As can be expected, Table \ref{table-results-summary} shows that results are similar to indoor conditions. The slight difference is due to non-idealities and the use of the same camera parameters as the indoor experiments. Event-based cameras still lack an automatic tuning system of their parameters that hopefully will be developed in a future generation of a camera.

\subsection{Face scale changes}
\begin{figure}[h]
	\centering
	\includegraphics[width=\columnwidth]{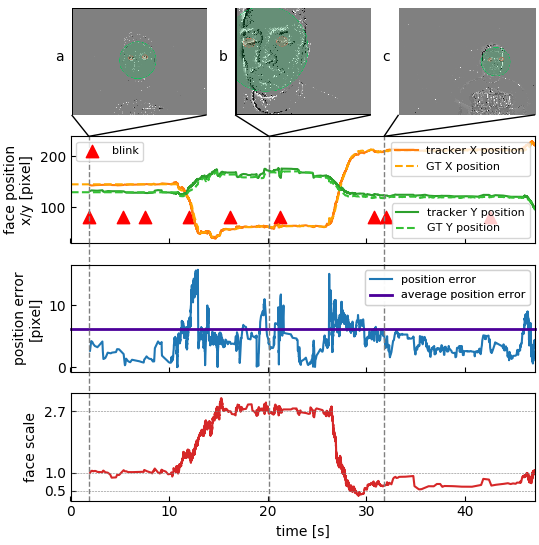}
	\caption{Verifying resistance to scale. a) first blink is detected at initial location. Scale value of 1 is assigned. b) Subject gets within 25cm of the camera, resulting in a three-fold scale change. c) Subject veers away to about 150cm, the face is now 35\% smaller than in a)}
	\label{fig:scale-data-recording}
\end{figure}

In 3 recordings the scale of a person's face varies by a factor of more than 5 from smallest to largest detected occurrence. Subjects sitting on a movable stool were instructed to approach the camera within 25\,cm after an initial position and then veer away again after 10\,s to about 150\,cm. Fig.~\ref{fig:scale-data-recording} shows tracking data for such a recording over time. The first blink is detected after 3\,s at roughly 1\,m in front of the camera (a). The subject then moves very close to the camera and to the left so that not even the whole face bounding box is seen anymore (b). Since the eyes are still visible, this is not a problem for our tracker. However, GT had to be partly manually annotated for this part of the recording, as two of the frame-based methods failed to detect the face that is too close to the camera. The subject then moves backwards and to the right, followed by further re-detections (c).


\subsection{Multiple faces detection}
\begin{figure}[h]
	\centering
	\includegraphics[width=\columnwidth]{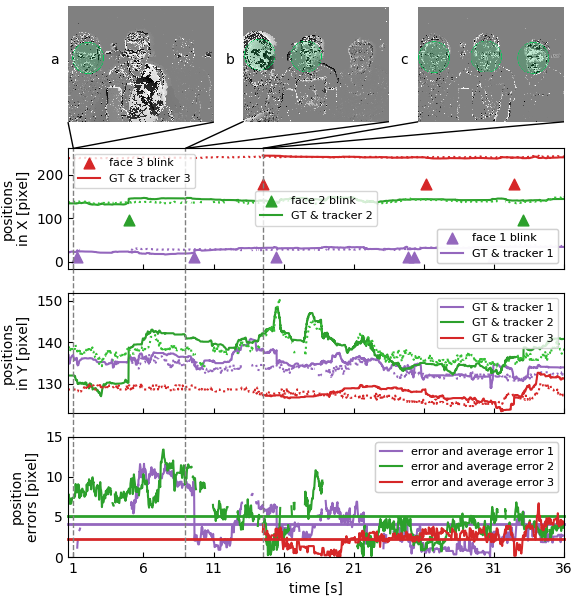}
	\caption{Multiple face tracking in parallel. Face positions in X and Y show three subjects sitting next to each other, their heads are roughly on the same height. a) subject to the left blinks at first. b) subject in the centre blinks next, considerably varying their face orientation when looking at the other two. c) third subject stays relatively still. }
	\label{fig:multiple-data-recording}
\end{figure}

In order to show that the algorithm can handle multiple faces at the same time, we recorded 3 sets of 3 subjects sitting at a desk talking to each other. No instructions where given, as the goal was to record in a natural environment. Fig.~\ref{fig:multiple-data-recording} shows tracking data for such a recording. The three subjects stay relatively still, but will look at each other from time to time as they are engaged in conversation or focus on a screen. Lower detection rates (see Table~\ref{table-results-summary}) are caused by an increased pose variation, however this does not result in an increase of the tracking errors.

\subsection{Occlusion sequences}
These last sequences aim to evaluate the robustness of the proposed method to pose variation that cause eye occlusions. The subjects in these sequences are rotating their head from one side to the other until one eye is partly occluded. Our experiments show that our algorithm successfully recovers from occlusion to track the eyes. These experiments have been carried out with an event-based camera at VGA resolution. While this camera provides better temporal accuracy and spatial resolution, it does not provide grey-level events measurements. Although we fed frames from the change detection events (which do not contain absolute grey-level information) to  the frame-based methods, none of them would sufficiently detect a face. This can be expected as the networks had been trained on grey-level images. We believe that if we retrained the last layers of the networks with manually labelled frames from change detection events, they would probably achieve similar performance. However the frame data set creation and the training are not the scope of this work.

\begin{figure*}[h]
	\centering
	\includegraphics[width=\linewidth]{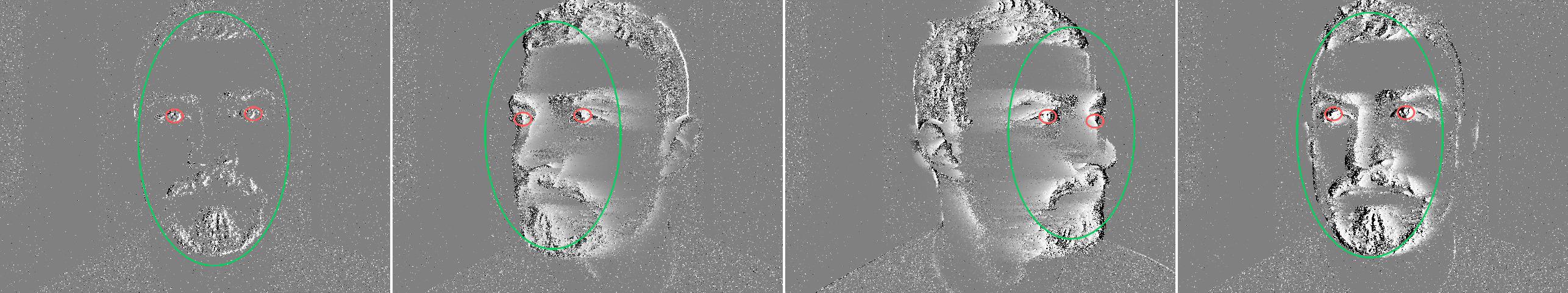}
	\caption{Pose variation experiment. \textbf{a}) Face tracker is initialised after blink. \textbf{b}) subject turns to the left. \textbf{c-d}) One eye is occluded, but tracker is able to recover.}
	\label{fig:pose-sequence}
\end{figure*}

\subsection{Summary}
Table~\ref{table-results-summary} summarises the accuracy of detection and tracking of the presented method, in comparison to Viola-Jones (VJ), Faster-RCNN (FRCNN) and Single Shot Detector (SSD) algorithms. The tracking errors are the deviations from the frame-based bounding box centre, normalised by the bounding box's width. The normalisation provides a scale invariance so that errors estimated for a large bounding box from a close-up face have the same meaning as errors for a small bounding box of a face further away.
\begin{table}
\renewcommand{\arraystretch}{1.3}
\centering
\begin{tabular}{p{1cm}|p{.8cm}|p{1.1cm}|p{0.8cm}|p{1.1cm}|p{0.9cm}}
& \small{\# of recordings} & \small{blinks detected (\%)}& \small{error VJ (\%)}& \small{error FRCNN (\%)}& \small{error SSD (\%)}\\ 
\hline
\small{indoor}  &   21 &     $68.4$            & $5.92$            &   $9.42$             &  $9.21$           \\
\small{outdoor} &   21 &     $52.3$            & $7.6$             &   $14.57$            & $15.08$           \\
\small{scale}   &    3 &     $62.6$            & $4.8$             &   $10.17$            & $10.22$           \\
\small{multiple}&    3 &     $36.8$            & $15$              &   $16.15$            & $14.61$           \\ 
\hline
\small{total}   &   48      &     $59$     & $7.68$      &   $11.77$    & $11.52$        \\
\end{tabular}
\caption{Summary of results for detection and tracking for 4 sets of experiments. \% of blinks detected relates to the total number of blinks in a recording. Tracking errors are Euclidean distances in pixel between the proposed and compared method's bounding boxes, normalised by the frame-based bounding box width and height.}
\label{table-results-summary}
\end{table}

\section{Conclusion}
The presented method for face detection and tracking is a novel method using an event-based formulation. It relies on eye blinks to detect and update the position of faces making use of dynamical properties of human faces rather than a approach which is purely spatial. 
The face's location is updated at $\mu$s precision that corresponds to the native temporal resolution of the camera. Tracking and re-detection are robust to more than a five-fold scale, corresponding to a distance in front of the camera ranging from 25\,cm to 1.50\,m. A blink seems to provide a sufficiently robust temporal signature as its overall duration changes little from subject to subject. 

The amount of events received and therefore the resulting activity amplitude varies only substantially when lighting of the scene is extremely different (i.e. indoor office lighting vs  bright outdoor sunlight). The model generated from an initial set of manually annotated blinks is proven robust to those changes across a wide set of sequences. Even so, we insist again in stating that the primary goal of this work is not to detect 100\,\% of blinks, but to reliably track a face. The blink detection acts as initialisation and recovery mechanism to allow that. This mechanism allows some resilience to eye occlusions when a face moves from side to side. In the most severe cases of occlusion, the tracker manages to reset correctly at the next detected blink.

The occlusion problem could be further mitigated by using additional trackers for more facial features (mouth, nose, etc) and by linking them to build a deformable part-based model of the face as it has been tested successfully in \cite{ReverterValeirasAsynchronousNeuromorphicEventDriven2015}. Once the trackers are initiated, they could more easily keep the same distances between parts of the face. This would also allow for a greater variety in pose variation and more specifically, this would allow us to handle conditions when subjects do not directly face the event-based camera.  

The blink detection approach is simple and yet robust enough for the technique to handle up to several faces simultaneously. We expect to be able to improve detection accuracy even more by learning the dynamics of blinks via techniques such as HOTS~\cite{LagorceHOTSHierarchyeventbased2016}. At the same time with increasingly efficient event-based cameras providing higher spatial resolution the algorithm is expected to increase its performances and range of operations. 
We roughly estimated the power consumption of the compared algorithms to provide numbers in terms of efficiency:
\begin{itemize}
    \item The presented event-based algorithm runs in real-time on 70\% of a single core of an Intel i5-7200U CPU for mobile Desktops, averaging to 5.5\,W of power consumption estimated from \cite{7thGenerationIntel2017}. 
    \item The OpenCV Viola Jones implementation is able to run 24 of the 25\,fps in real-time, using one full core at 7.5\,W again inferred from 15W full load for both cores\cite{7thGenerationIntel2017}.
    \item The Faster R-CNN Caffe implementation running on the GPU uses 175\,W on average on a Nvidia Tesla K40c with 4-5\,fps. 
    \item The SSD implementation in Tensorflow runs in real-time, using 106\,W on average on the same GPU model. 
\end{itemize}

Currently our implementation runs on a single CPU core, beating SOA neural nets by an estimated factor of 20 in terms of power efficiency. Due to the asynchronous nature of the input and our method that adapts to it, it could easily be parallelised across multiple threads, using current architecture that is still bound to synchronous processing of instructions and allocation of memory. A neural network model that runs on neuromorphic hardware could further improve power efficiency by a factor of at least 10.

{\small
\bibliographystyle{ieee}

}

\end{document}